# YOIO: You Only Iterate Once by mining and fusing multiple necessary global information in the optical flow estimation

Yu Jing[1], Tan Yujuan[1], Ren Ao[1], Liu Duo[1]
[1]Chongqing University
{20201401010, tanyujuan, ren.ao, liuduo}@cqu.edu.cn

## Abstract

*Occlusions pose a significant challenge to optical flow algorithms that even rely on global evidences. We consider an occluded point to be one that is imaged in the reference frame but not in the next. Estimating the motion of these points is extremely difficult, particularly in the two-frame setting. Previous work only used the current frame as the only input, which could not guarantee providing correct global reference information for occluded points, and had problems such as long calculation time and poor accuracy in predicting optical flow at occluded points. To enable both high accuracy and efficiency, We fully mine and utilize the spatiotemporal information provided by the frame pair, design a loopback judgment algorithm to ensure that correct global reference information is obtained, mine multiple necessary global information, and design an efficient refinement module that fuses these global information. Specifically, we propose a YOIO framework, which consists of three main components: an initial flow estimator, a multiple global information extraction module, and a unified refinement module. We demonstrate that optical flow estimates in the occluded regions can be significantly improved in only one iteration without damaging the performance in non-occluded regions. Compared with GMA, the optical flow prediction accuracy of this method in the occluded area is improved by more than 10%, and the occ_out area exceeds 15%, while the calculation time is 27% shorter. This approach, running up to 18.9fps with 436\*1024 image resolution, obtains new state-of-the-art results on the challenging Sintel dataset among all published and unpublished approaches that can run in real-time, suggesting a new paradigm for accurate and efficient optical flow estimation.*

## 1. Introduction

There are many factors that make optical flow prediction a hard problem, including large motions, motion and defocus blur, and featureless regions. occlusion is one of the most difficult Among these challenges. We first define what we mean by occlusion in the context of optical flow estimation same as GMA [6], i.e., an occluded point is defined as a 3D point that is imaged in the reference frame but is not visible in the matching frame. This definition incorporates several different scenarios, such as the query point moving out-of-frame(occ_out) or behind another object(or itself), or another object moving in front of the query point, in the active sense(occ_in). Figure 1 provides some examples of occ_out. One definition of optical flow is the coordinate difference between pixel points projected by the same thing in two images. Therefore, because the occluded (occ) point is not in the next frame, the pixel coordinates of the occluded point in the next frame cannot be obtained directly through search or matching. This is one of the important reasons why optical flow at occluded point is difficult to estimate. At present, optical flow at occluded point is mainly estimated through indirect means [5, 6, 8, 20-25]. The main way is to first find the non-occluded (noc) point that is contextually correlated (also called self-similar) to the occluded point, then use the optical flow of the non-occluded point to reasonably guess the optical flow of the correlated occluded point. However, at present, the error of optical flow at occluded points using these methods is still very large. In this paper, we propose an approach that specifically targets the occlusion problem in the case of two-frame optical flow prediction.

In order to solve the problem of large optical flow prediction error at occluded points, GMA [6] proposed a refinement method for optical flow at occlusion points based on the self-similarity assumption. The self-similarity assumption holds that the motions of a single object (in the foreground or background) are often homogeneous, so the optical flow of the non-occluded points can be used to refine the optical flow of the occluded points on the surface of the same object. The necessary processes and rules for designing a method based on this assumption to refine optical flow at the occluded point can be deduced: The first step is to find the reference point set P_ref and the set to be refined P_pre, as shown in figure 1(c). It is necessary to ensure that P_ref is a subset of all non-occluded points because only non-occluded points can provide correct reference information. An occ point can only belong to P_pre. P_pre must at least contain all occluded points that need to be refined. P_pre could also include non-occluded points because, in principle, non-occluded points can also be refined correctly using reference information. The second step is to find the specific reference points corresponding to each point to be refined to form a global correlation pair: <p_ref,p_pre>, p_ref comes from P_ref, p_pre comes from P_pre, p_ref provides reference information for p_pre. The third step is to find a reasonable way to extract enough reference information from the global correlation pair to refine the optical flow of p_pre perfectly.

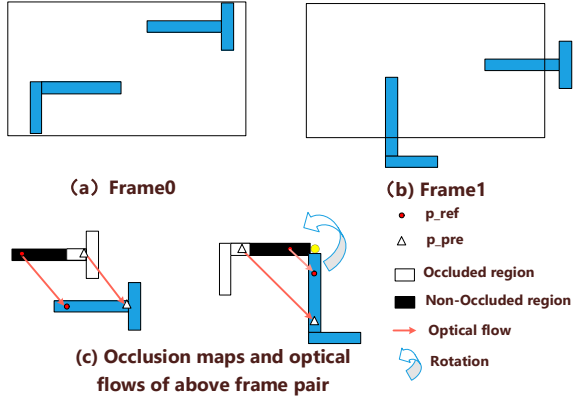

Figure 1: (a) and (b) form a frame pair. In frame0, there are two blue objects, L-shaped and T-shaped. (c) is the corresponding occlusion maps and optical flows of the above frame pair. An occlusion map is the sum of all non-occluded regions and non-occluded regions. If we know all of the non-occluded regions in frame0, the rest of the regions of frame0 are all occluded regions.

The current methods based on the self-similarity assumption mainly include GMA [6] and GMFlow [8]. They all have a common flaw that leads to poor refinement results. They cannot ensure that the reference set P_ref is a subset of all non-occluded points. This is because they obtain P_ref by performing self-attention on frame0, but a single frame does not have enough information to determine whether a point to be p_ref is occluded or not. Since occlusions are caused by different times (shooting at different times) or different spaces (different shooting angles), this spatiotemporal information is also necessary to determine whether a point is occluded or not, and a single frame lacks this information. Because they cannot ensure providing reliable reference information, refinement performance based on unreliable reference information is poor.

Currently there has some related works to find ways of determining whether a point is occluded, and they all use frame pairs as input. However, they can only obtain the occlusion map and not the global correlation pair, resulting in the inability to use these methods to provide reference information for the occluded points.

Therefore, key problem 1 that still remains to be solved is: how to obtain reliable P_ref and global correlation pair which provides correct reference information for the subsequent refinement processes to behave well.

Assuming that the above two steps have been performed correctly, now it is time to perform the third step. The implementations of GMA and GMFlow [8] have different characteristics, as shown in Figure 2(b). GMFlow [8] directly takes the optical flow of p_ref in the global correlation pair and gives it to the corresponding p_pre. The advantage of the above way is that it reduces the computational complexity of the method, but sacrifices the refinement accuracy. When points in the global correlation pair perform rotation or even more complex motion together, the refinement result of GMFlow [8] has a larger error. Because when the object is in rotational motion, the difference in optical flow between points on the same object's surface may increase as the distance between them increases, as shown in the rotating L-shaped object in Figure 1(c). GMA is a multiple-iteration method based on RAFT [5]. RAFT can implicitly determine whether it is an occlusion point through the local cost volume, use the context information to associate the non-occluded and occluded points, and then refine the optical flow of the occluded point based on the optical flow of its correlated non-occluded point. However, since RAFT only has a local field of view in each iteration, it lacks the global information to indicate the correct correction direction when the local field of view is full of occluded points. Therefore, GMA extracts and fuses global motion information to make up for this shortcoming, and has made certain progress. But GMA still requires a sufficient number of iterations to ensure that all the information of the non-occluded point is transferred to the occluded point far away from it, which causes the required calculation time to increase linearly with the number of iterations, as shown in 2(b).

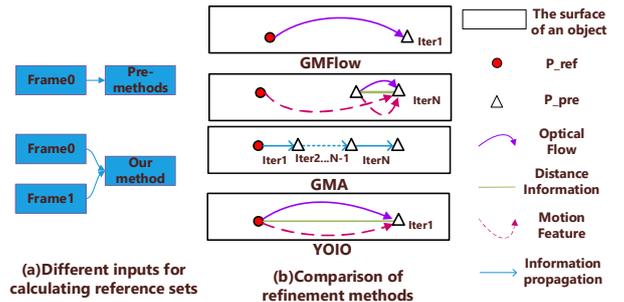

Figure 2: in (a), Different inputs for calculating reference sets, and in (b), Comparison of refinement methods. "Iter" is the abbreviation of iteration. "Iter N" represents the nth iteration. An iteration is a refinement.

Therefore, the key issue 2 that still remains to be solved is: how to efficiently obtain the accurate optical flow of all p_pre points through refinement.

In order to solve the first problem, we are based on the principle that the point is most similar to the point itself in the global context, and the self-similar points other than the point itself are the second most similar to it. We use frame pairs as the input of the multiple global information extraction module, then in it, a loopback judgment algorithm is designed to find the correct reference set, and at the same time construct the correct global correlation pair, and finally extract a variety of necessary global information based on the global correlation pair.

In order to solve the second problem, we design an efficient refinement module that fuses multiple global information, and the performances of one refinement are significantly better than GMA.

The key contributions of our paper are as follows.

We have designed multiple global information extraction modules. In this module, a loopback judgment algorithm based on frame pairs is designed to ensure that correct global information is obtained.

Our method can directly predict the occlusion map without backward flow computation. An important highlight is that we also do not need occlusion map labels to train our network.

We proposed and designed a refinement module that fuses multiple global information, which significantly improves computational efficiency and accuracy.
Compared with GMA, the optical flow refinement accuracy of this method in the occluded area is improved by more than 10%, the occ_out area exceeds 15%, and the calculation time is 27% shorter, suggesting a new paradigm for accurate and efficient flow estimation.

## 2. Related work

**Flow estimation approach.** The flow estimation approach is fundamental to existing popular optical flow frameworks [3, 5, 6, 16, 17-19], notably the coarse-to-fine method PWC-Net [17] and the iterative refinement method RAFT [5]. They both perform some sort of multi-stage refinements, either at multiple scales [17] or a single resolution [5]. For flow prediction at each stage, their pipeline is conceptually similar, i.e., regressing optical flow from a local cost volume with convolutions. Thus multi-stage refinements are required to estimate large motion incrementally. There will be a lot of follow-up work to improve various aspects under the RAFT architecture, using the transformer to enhance the feature map extracted by CNN, so as to better deal with difficult problems such as less texture, motion blur, and large motion [20-24]. In order to improve the optical flow estimation effect in occlusion areas, GMA [6] proposed a global motion aggregation module. In order to overcome the problem that multiple iterations require a large amount of memory when training the network and the optimization results are unstable, RAFT based on DEQ [25] was proposed. The success of these RAFT-like methods largely lies in the large number of iterative refinements they can perform, which results in long calculation times and is difficult to apply in practical applications. To enable both high accuracy and efficiency, GMFlow [8] completely revamps the dominant flow regression pipeline by reformulating optical flow as a global matching problem, which identifies the correspondences by directly comparing feature similarities. They all have the problem of insufficient exploration of the optical flow estimation problem at the occlusion point, resulting in poor optical flow estimation results at the occlusion point.

**Occlusions in optical flow.** The main problem with these methods currently is that there is a large error in the optical flow at the occlusion point, especially an order of magnitude larger than the average error at the non-occluded points [6, 8, 20-24]. Optical flow estimation at occlusion points is divided into two major categories of methods. The first type of method is to explicitly determine whether it is an occlusion point [1, 2, 26-30], and the second type of method is to implicitly determine whether it is an occlusion point [1, 2, 26-30]. 3-8, 16, 17-19, 20-25], both methods hope to know and use the occlusion map information to help them improve the optical flow estimation effect of occlusion points.

The first type of method can be further divided into two subcategories. The first subcategory is to directly predict the occlusion map as output[28-30], and the second subcategory is based on the forward-backward constraint assumption [1, 2, 26, 27]. Methods that directly predict occlusion maps require occlusion labels for network training. Since occlusion labels are difficult to produce, there are few data sets available for training. The method based on the forward-backward constraint assumption requires the calculation of bidirectional optical flow, resulting in too long calculation time. Since only the occlusion map is obtained and there is a lack of other necessary reference information, the effect of optical flow prediction for improving the occlusion point is not well.

RAFT [5] belongs to the second type of method. It can implicitly determine whether it is an occlusion point through the local cost volume, associate non-occluded points and occlusion points with context information, and then use the optical flow of the associated non-occluded points to predict the optical flow of occlusion points. However, since RAFT [5] only has a local field of view in each iteration. When the local field of view is full of occlusion points, there is a lack of necessary global information to indicate the correct correction direction, resulting in poor results in this case. Therefore, GMA [6] extracts and fuses global motion information to make up for this shortcoming, and has made certain progress. However, it still requires a sufficient number of iterations to ensure that all the information of the non-occluded point is transferred to the correlated occluded point far away from it, which causes the required calculation time to increase linearly with the number of iterations. Another method of using global information to predict the optical flow of occlusion points is GMFlow, but it directly takes the optical flow of the reference point and gives it to the point to be predicted. The advantage of this is that it reduces the computational complexity of the method, but it sacrifices the prediction accuracy. Therefore, GMFlow performs poorly in complex motion scenarios. There is another serious problem with GMA [6] and GMFlow [8]. They cannot guarantee to obtain the correct global reference information with only a single frame. In contrast, We use frame pairs as input, design a loopback judgment algorithm to ensure correct reference information, extract and fuse multiple necessary global information to reduce the number of iterations to only once, and achieve better results at the same time.

## 3. Methodology

### 3.1. Background

We calculate the local and global feature maps, as well as the initial optical flow Flow0 based on the method of GMFlow [8], as shown in figure 3.

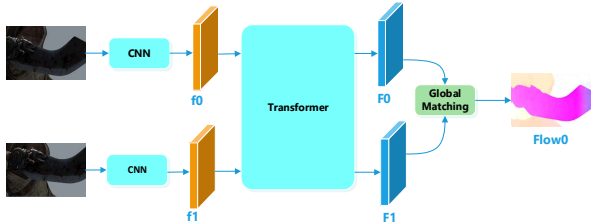

Figure 3: The overall process of calculating Flow0. GMFlow first extracts 8× downsampled dense features from two input video frames with a weight-sharing convolutional network. Then the features are fed into a Transformer for feature enhancement. Next, it compares feature similarities by correlating all pair-wise features and the optical flow is obtained with a softmax matching layer.

### 3.2. Overview

We first design the loopback judgment algorithm to find a reliable reference point set, and at the same time build all global correlation pairs, and then extract a variety of global reference information, see Sec. 3.3 for details. Since occ_in points are special and need to be processed separately, we find a way to extract an advanced feature map that can identify occ_in points. See Sec. 3.4 for details. In order to further improve the optical flow accuracy of non-occluded points, we build the local cost volume like RAFT [5], see Sec. 3.5 for details. Finally, we splice the information extracted in all the above sections and send it to the unified refinement module, so that it can automatically identify various types of points and refine their optical flow properly, see Sec. 3.6 for details.

### 3.3. Multiple Global Information Extraction Module

In Flow0, the optical flow of the non-occluded (noc) points is accurate, but the optical flow of the occluded (occ) points is less accurate [8], so we must refine the optical flow of these occ points. To refine according to the self-similarity principle, we need to first find the reference point p_ref that the occ point can refer to. Because only the optical flow of noc in Flow0 is accurate, only noc points can be used as p_ref points [8]. But at this time we don't know which are noc points and which are occ points in frame0.

**The loopback judgment algorithm to find a reliable reference point set.** In order to find a p_ref point and ensure that the found p_ref is a noc point, we designed a loopback judgment algorithm. We first introduce the principle that in the global context, the point is most similar to the point itself in any frame (principle 1), and the self-similar points other than the point itself are the second most similar to it (principle 2). These principles generally hold true in matching methods based on global features [32-34]. Then the algorithm process is as follows: Based on the global features, for any point p0 in frame0, we find the point p1 that is most similar to it in frame1; in turn, we find the point p0' that is most similar to p1 from frame0. If p0 and p0' are the same point (a loopback occurs), based on principle 1, p0 must be a noc point. This can be proved by contradiction. Assuming p0 is a noc point, then p0 exists in frame1. A point p1' will be found from frame1, and p1' will find the most similar point p0' from frame0. At this time, p0' and p0 are still the same point according to principle 1. And if p0 and p0' are not the same point, it can be judged that p0 is an occ point.

According to principle 2, When p0 is an occ point, the found p1 is its self-similar point. There is a high probability that p0' is a non-occlusion point, only in rare cases p0' is an occ point, such as when the objects to which p0 and p0' belong do not exist in frame1 at all. Therefore, p0' can be used as the reference point of p0 to form a global correlation pair <p0', p0>, regardless of whether p0 is an occ point or a noc point.

The specific execution operation is shown in figure 4. All p1 can be obtained with:

$$\text{Flow0} + \text{Init\_coord} \quad (1)$$

Init_coord can be obtained by using the library function in torch [35], i.e.,

$$\text{torch.meshgrid}(\text{torch.arange}(h), \text{torch.arange}(w)) \quad (2)$$

where h and w are the height and width of the feature map respectively. Then use p1 to linear sample F1, construct a new F1', and then use F1' and F0 to perform global matching to obtain Flow_ref. Use

$$\text{Flow\_ref} + \text{init\_coord} \quad (3)$$

to find the corresponding reference point p0' for each point p0, so that the global correlation pair can be obtained.

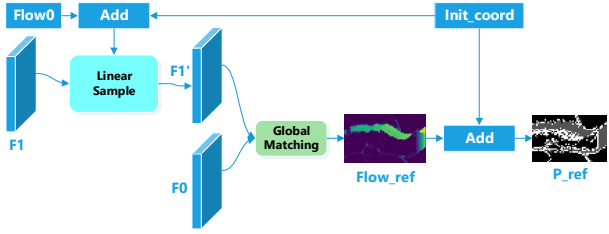

Figure 4: the loopback judgment algorithm to find a reliable reference point set.

**Occlusion map.** Whether p0 is occluded can be determined by the norm of Flow_ref at its pixel coordinate. When the norm is 0, p0 and p0' coincide, p0 is a non-occlusion point, otherwise, it is an occluded point.

Since the global features and matching methods used to calculate Flow_ref and Flow0 are the same, this method only needs to use optical flow labels for supervised learning without occlusion labels. And it is more resource-saving than the bidirectional optical flow method to calculate the occlusion map.

GMFlow accelerates bidirectional optical flow calculations by directly transposing the global correlation matrix to quickly obtain the occlusion map, but it still requires a lot of calculations, that is,

$$3G + \text{softmax}(\text{4D matrix}) + \text{additional} \quad (4)$$

while our method only requires 2G, where G stands for Global Matching, 4D matrix has a dimension of h × w × h × w.

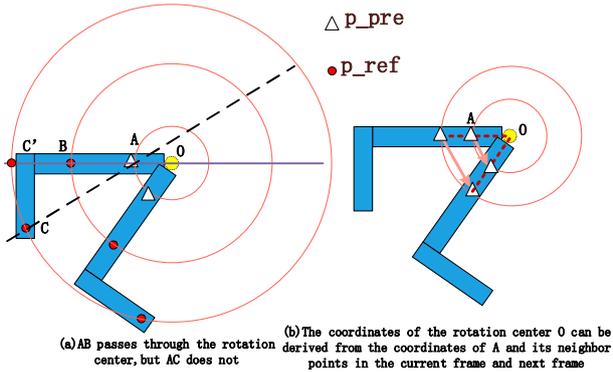

(a) AB passes through the rotation center, but AC does not
(b) The coordinates of the rotation center O can be derived from the coordinates of A and its neighbor points in the current frame and next frame

Figure 5: (a): AB passes through the rotation center, but AC does not. (b): The coordinates of the rotation center O can be derived from the coordinates of A and its neighbors in the current frame and next frame.

Similarly, we use Flow_ref to get the global reference optical flow Ref-flow from Flow0.

**The distance that conforms to the uniform law.** Because the object is in rotational motion, the difference in optical flow between points on the same object's surface may increase as the distance between them increases, as shown in the rotating L-shaped object in Figure 1(c). Therefore we need to estimate the distance between p_pre and p_ref. Simply taking the Euclidean distance between p_pre and p_ref as the distance may not be appropriate under rotational motion. As shown in figure 5, assuming that B and C both take A as the reference point, the straight line AB passes through the rotation center, and the straight line AC does not pass through the rotation center. In this case, The length difference between AB and BC is not uniformly related to their optical flow difference. Based on Euclidean distance does not follow a uniform law in this case.

Therefore, we can use the distance from the point to the rotation center O as the distance, such as:

$$\text{torch.abs}(|AB| - |AO|) \quad (5)$$

In order to calculate the coordinates of the rotation center, we can use the coordinates of the reference point A and its neighbors in the current frame and next frame. We can use Flow_ref to sample that information, input it into a convolutional network, and output the rotation center coordinates.

### 3.4. Extract the High-level Feature Map that Identifies Occ_in Points

When Flow0 is added to the loss function for loss calculation, in order to minimize the loss, the occ_in point will match the non-occluded point that occludes it. This is reasonable because cross-attention can learn to recognize such globally relevant information [8]. Therefore, they have very similar global features, the optical flow of the occ_in point is already accurate in Flow0, and they form a global correlation pair. However, occ_in points and correlated non-occluded points often do not come from the same object or segment and do not satisfy the self-similarity assumption, so the global correlation pair formed by them is incorrect. Therefore, the occ_in points need to be identified and processed separately.

Since occ_in points and correlated non-occluded points come from different objects or segments, their local features are different. Therefore, they have the characteristics that global features are similar to each other, but local features are dissimilar. It can be seen from principle 2 mentioned in section 3.3. Other global correlation pairs come from the same object or segment, and their global features are similar and their local features are also similar. So we can use this difference to identify occ_in points.

The specific operation is that: We first obtain the local feature similarity map through the dot product of f0 and f1 and obtain the global feature similarity map through the dot product of F0 and F1, and then splice these two maps and send them to a convolution network to extract more advanced feature maps which providing necessary information for identifying occ_in points. The flowchart of this operation is shown in the supplementary material.

### 3.5. Build Local Cost Volume

Inspired by RAFT, we use the local regression method to further improve the optical flow accuracy of noc points. We use Flow0 as the initial optical flow and construct a cost volume based on the local feature map. Since the average end point error of optical flow in the noc area of GMFLOW is less than 1, we set the correlation radius to 3, which can reasonably save computing resource.

### 3.6. Unified Refinement Module

we splice the information extracted in all the above sections and send it to the unified refinement module so that we can train it to learn how to automatically identify various types of points and refine the optical flow of them properly. We implement this module using a convolutional network.

We let the network output weight and residual, and use

$$(1 - weight) * Flow0 + weight * residual \qquad (6)$$

to predict the final result. When weight=0, only Flow0 is left, that is, no correction is made to this point, which corresponds to the occ_in points.

We supervise all flow predictions like GMFlow [8].

Flowchart and more detail results of this module are shown in the supplementary material.

## 4. Experiments

### 4.1. Experimental Setup

**Datasets and evaluation setup.** Following previous methods [3, 5, 8, 17], we first train on the Flying-Chairs(Chairs) [15] and FlyingThings3D (Things) [13] datasets, and then evaluate Sintel [14] and KITTI [11] training sets. Finally, we perform additional fine-tuning on Sintel training sets and report the performance on the online benchmarks.

**Metrics.** We adopt the commonly used metric in optical flow, i.e., the average end-point-error (AEPE), which is the average $\ell 2$ distance between the prediction and ground truth.

**Implementation details.** We implement our framework in PyTorch. Our feature extraction backbone network is identical to GMFlow model. We also stack 6 Transformer blocks. To upsample the low-resolution flow prediction to the original image resolution, we use RAFT's convex upsampling [5] method. We use AdamW [25] as the optimizer. We, first train the model on the Chairs dataset for 100K iterations, with a batch size of 16 and a learning rate of 4e-4. We then fine-tune it on the Things dataset for 800K iterations for ablation experiments, with a batch size of 8 and a learning rate of 2e-4. For the final fine-tuning process on Sintel datasets, we further fine-tune our Things model on several mixed datasets that consist of KITTI [11], HD1K [12], FlyingThings3D [13], and Sintel [14] training sets. We perform fine-tuning for 200K iterations with a batch size of 8 and a learning rate of 2e-4.

### 4.2. Comparison with GMFlow

Since there is no mechanism to ensure that the reference set is correct, and refinement is based on imperfect models, the error in the noc area in GMFlow Flow0 is large. To verify that, we used the network weights trained on the Things data set and conducted experiments on the clean subset of the Sintel training data set. The experimental setup is in section 4.1. GMFlow 1/8 and our YOIO all only use a feature map with a resolution of 1/8 of the frame0.

| Sintel Pass | Type | GMFlow 1/8 Flow0 (AEPE) | YOIO Flow0 (AEPE) | Rel.Impr. (%) |
|---|---|---|---|---|
| Clean | Noc | 0.89 | **0.69** | 22.47191 |

Table 1: Quantitative results on Sintel clean sub-datasets in non-occluded ('Noc') regions. We report the average end-point error (AEPE) where not otherwise stated.

| Sintel Pass | Type | GMFlow 1/8 Flow0 (AEPE) | YOIO Flow0 (AEPE) | Rel.Impr. (%) |
|---|---|---|---|---|
| Clean | Noc | 0.81 | **0.69** | 22.47191 |

Table 2: Quantitative results on Sintel clean sub-datasets.

Table 1 shows the AEPE of Flow0 output by these two methods in the non-occluded (noc) area. Table 2 shows the AEPE of flow1 output by GMFlow 1/8 and Flow0 output by our method in the noc area. Obviously, In the noc area, the accuracy of Flow0 and flow1 output by GMFlow 1/8 is obviously not as good as that of our method. This is indicating that our method is more reasonable which first ensures the correctness of the reference set, and then extracts and fuses multiple necessary global information to do refinement.

We visually demonstrate the performance differences between the two methods under rotational motion, as shown in figure 6. The first row is frame0 and frame1 respectively. The things contained in the red box have made a rotational movement with the tail of the knife as the center of rotation. The second row from left to right is the optical flow output from GMFlow and YOIO respectively, and the third row is the EPE map corresponding to the optical flow of the previous row. In the EPE map, the brighter the color, the greater the error. It can be clearly seen from the EPE map of GMFlow that the farther away from the rotation center, the greater the error. From the predicted optical flow of GMFlow, it can be seen that there is almost no change in the optical flow in the red box area. This indicates that GMFlow directly takes the optical flow value of the reference point and gives it to its correlated occluded point without considering other necessary global information, causing it to perform poorly under rotational motion, and the greater the distance between the reference point and its correlated occluded point, the worse the performance. On the contrary, our method performs very well. This shows that our approach is more reasonable and fully exploring and utilizing the relationship between reference points and correlated occluded points is beneficial to improving refinement accuracy.

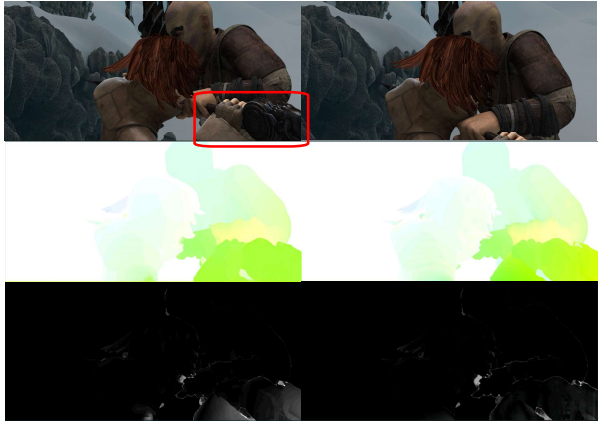

Figure 6: The first row is the frame pair, the second row from left to right is the optical flow output from GMFlow and YOIO respectively, and the third row is the EPE map corresponding to the optical flow of the previous row. In the EPE map, the brighter the color, the greater the error.

### 4.3. Comparison with GMA

The setup used in this experiment is the same as in section 4.2. In table 3, what is more interesting is that all the results have the same characteristic, that is, the EPE of GMA in the non-occluded region is better than our method, but in the occluded area is worse.

We think the main reason why GMA performs better in the non-occluded region is that GMA uses multiple iterative local refinements based on local cost volume to achieve higher accuracy, and we only use one refinement.

As shown in figure 7, our method performs much better in occluded regions after one refinement than GMA after 12 or more iterative refinement. Compared with GMA, the optical flow prediction accuracy of our method in the occluded area is improved by more than 10%, and the out-of-frame occlusions exceed 15%, while the calculation time is 27% shorter (53ms vs 72ms, NVIDIA RTX 3090, 4, 436*1024 image resolution). This shows two points: (1)Since GMA cannot guarantee the correctness of the reference set and provides incorrect global reference information, the error in occluded regions is large; (2) On the contrary, ensuring the correctness of the reference set, providing correct reference information, and extracting and fusing a variety of global information make our method better and faster.

| Sintel Pass | Type | GMA (AEPE) | YOIO (AEPE) | Rel.Impr. (%) |
|---|---|---|---|---|
| Clean | Noc | **0.58** | 0.67 | -15.5172 |
| | Occ | 10.58 | **9.51** | 10.11342 |
| | Occ-in | 7.68 | **7.3** | 4.947917 |
| | Occ-out | 12.52 | **10.62** | 15.17572 |
| | All | **1.3** | 1.32 | -1.53846 |
| Final | Noc | **1.72** | 1.73 | -0.5814 |
| | Occ | 17.33 | **15.86** | 8.4824 |
| | Occ-in | 14.96 | **13.55** | 9.425134 |
| | Occ-out | 16.44 | **15.33** | 6.751825 |
| | All | **2.74** | 2.77 | -1.09489 |
| Albedo | Noc | **0.48** | 0.55 | -14.5833 |
| | Occ | 9.54 | **8.8** | 7.756813 |
| | Occ-in | 7.51 | **7.22** | 3.861518 |
| | Occ-out | 10.22 | **8.94** | 12.52446 |
| | All | 1.15 | 1.15 | 0 |

Table 3: Optical flow error for different Sintel datasets, partitioned into occluded ('Occ') and non-occluded ('Noc') regions. In-frame and out-of-frame occlusions are further split and denoted as 'Occ-in' and 'Occ-out'. The best results and the largest relative improvement in each dataset are styled in bold.

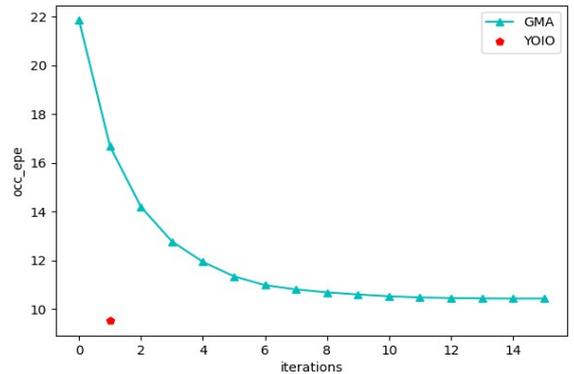

Figure 7: Comparison of EPE in occluded regions (occ_EPE) between GMA and YOIO. The abscissa axis represents occ_EPE, and the ordinate axis represents the number of iterations.

### 4.4. Qualitative Results

The experimental setup is Sec. 4.1. All experimental results are in table 4. In the generalization verification experiment (C+T), GMFlow 1/8 did not surpass GMA and our method. GMFlow 1/4 further performs local refinement based on a feature map with a resolution of 1/4 of the frame0, making it surpass the previous two methods. There is an interesting phenomenon in these experimental results. Although GMFlow 1/8 does not exceed GMA in the non-occluded regions, exceeds GMA in the occluded regions (do not exceed our method). We think a reasonable explanation is as follows:

The Sintel train data set contains more translational movements. So in this case, there is nothing wrong with GMFlow directly taking the optical flow of the reference point and giving it to the occ point. But because of this, GMFlow performs worse performance on the Sintel test, because there are more complex movement patterns and more strenuous movements [14]. This shows that the refinement method designed by GMFlow is imperfect.

| Training Stage | Method | Sintel (clean) | | |
|---|---|---|---|---|
| | | all | matched | unmatched |
| C+T | GMA [6] | 1.3 | 0.58 | 10.58 |
| | GMFlow 1/8 [8] | 1.5 | 0.86 | 9.8 |
| | GMFlow 1/4 [8] | **1.08** | **0.51** | **8.27** |
| | YOIO | 1.35 | 0.67 | 9.51 |
| C+T+S+K+H | FlowNet 2 [3] | 4.16 | 1.56 | 25.4 |
| | PWC-Net+ [4] | 3.45 | 1.41 | 20.12 |
| | HD3 [9] | 4.79 | 1.62 | 30.63 |
| | VCN [7] | 2.81 | 1.11 | 16.68 |
| | DICL [10] | 2.63 | 0.97 | 16.24 |
| | RAFT [5] | 1.94 | - | - |
| | GMA [6] | 1.391 | **0.562** | 8.137 |
| | GMFlow 1/4 [8] | 1.74 | 0.65 | 10.56 |
| | YOIO | **1.365** | 0.599 | **7.584** |

Table 4: Quantitative results on Sintel datasets. "C + T" refers to results that are pre-trained on the Chairs and Things datasets. "S + K + H" refers to methods that are fine-tuned on the Sintel and KITTI datasets, with some also fine-tuned on the HD1K dataset. Parentheses denote training set results and bold font denotes the best result.

Our method hold the same performance on the test data set, and outperforms both GMA and GMFlow, which fully show that our method is more reasonable and effective.

In order to save pages, we put more experimental results into supplementary material.

### 4.5. Ablation Results

To verify our design, we conducted the following ablation experiments. All experimental results are in table 5.

We use whether to add distance information to reflect the rationality of our method. We can see that the performance in occluded regions gradually increases by adding and using more reasonable distance. Our method performs best with distance that confirms the uniform law, which shows that it is necessary to fuse distance information, and using distance that confirms the uniform law is more helpful to improve the performance.

| | Sintel (train clean) |
|---|---|
| Component | Occ AEPE |
| without distance | 9.69 |
| with Euclidean distance between p_pre and p_ref | 9.55 |
| with distance that confirms the uniform law | **9.51** |

Table 5. Ablation experiment results.

### 4.6. Limitation and Discussion

Since the global feature extraction module we rely on is the same as GMFlow [8], therefore our method is still not generalized very well when the training data has a significantly large gap with the test data (e.g., synthetic Things to real-world KITTI). More data sets are needed to enhance Transformer's generalization ability.

In addition, the optical flow accuracy of our method in non-occluded areas still has a lot of room for improvement. Given a relatively accurate initial optical flow and setting a suitable search radius, why are the results of a single refinement worse than those of multiple iterations of refinement? This is a very interesting question.

## 5. Conclusion

Occlusions have long been considered a significant challenge and a major source of error in optical flow estimation. We have presented a new way of obtaining multiple global information and a new single refinement method, and demonstrated its strong performance. We hope our new perspective will pave the way towards a new paradigm for accurate and efficient optical flow estimation. Subsequent work will be dedicated to improving the optical flow accuracy in non-occluded areas based on local refinement that only iterates once.

## References

[1] Luis Alvarez, Rachid Deriche, Th´eo Papadopoulo, and Javier S´anchez. Symmetrical dense optical flow estimation with occlusions detection. IJCV, 2007.

[2] Junhwa Hur and Stefan Roth. Mirrorflow: Exploiting symmetries in joint optical flow and occlusion estimation. ICCV, 2017.

[3] Ilg E, Mayer N, Saikia T, et al. Flownet 2.0: Evolution of optical flow estimation with deep networks[C]//Proceedings of the IEEE conference on computer vision and pattern recognition. 2017: 2462-2470.

[4] Deqing Sun, Xiaodong Yang, Ming-Yu Liu, and Jan Kautz. Models matter, so does training: An empirical study of cnns for optical flow estimation. TPAMI, 42(6):1408–1423, 2019.

[5] Zachary Teed and Jia Deng. RAFT: Recurrent all-pairs field transforms for optical flow. ECCV, 2020.

[6] Jiang, Shihao, et al. "Learning to estimate hidden motions with global motion aggregation." Proceedings of the IEEE/CVF International Conference on Computer Vision. 2021.

[7] Gengshan Yang and Deva Ramanan. Volumetric correspondence networks for optical flow. NeurIPS, 32:794–805, 2019.

[8] Xu, Haofei, et al. "GMFlow: Learning optical flow via global matching." Proceedings of the IEEE/CVF conference on computer vision and pattern recognition. 2022.

[9] Zhichao Yin, Trevor Darrell, and Fisher Yu. Hierarchical discrete distribution decomposition for match density estimation. In CVPR, pages 6044–6053, 2019.

[10] Jianyuan Wang, Yiran Zhong, Yuchao Dai, Kaihao Zhang, Pan Ji, and Hongdong Li. Displacement-invariant matching cost learning for accurate optical flow estimation. NeurIPS, 33, 2020.

[11] Moritz Menze and Andreas Geiger. Object scene flow for autonomous vehicles. In CVPR, pages 3061–3070, 2015.

[12] Daniel Kondermann, Rahul Nair, Katrin Honauer, Karsten Krispin, Jonas Andrulis, Alexander Brock, Burkhard Gussefeld, Mohsen Rahimimoghaddam, Sabine Hofmann, Claus Brenner, et al. The hci benchmark suite: Stereo and flow ground truth with uncertainties for urban autonomous driving. In CVPR Workshops, pages 19–28, 2016.

[13] Nikolaus Mayer, Eddy Ilg, Philip Hausser, Philipp Fischer, Daniel Cremers, Alexey Dosovitskiy, and Thomas Brox. A large dataset to train convolutional networks for disparity, optical flow, and scene flow estimation. In CVPR, pages 4040–4048, 2016.

[14] Daniel J Butler, Jonas Wulff, Garrett B Stanley, and Michael J Black. A naturalistic open source movie for op- tical flow evaluation. In ECCV, pages 611–625. Springer, 2012.

[15] Alexey Dosovitskiy, Philipp Fischer, Eddy Ilg, Philip Hausser, Caner Hazirbas, Vladimir Golkov, Patrick Van Der Smagt, Daniel Cremers, and Thomas Brox. Flownet: Learning optical flow with convolutional networks. In ICCV, pages 2758–2766, 2015.

[16] Junhwa Hur and Stefan Roth. Iterative residual refinement for joint optical flow and occlusion estimation. In CVPR, pages 5754–5763, 2019.

[17] Deqing Sun, Xiaodong Yang, Ming-Yu Liu, and Jan Kautz. Pwc-net: Cnns for optical flow using pyramid, warping, and cost volume. In CVPR, pages 8934–8943, 2018.

[18] Haofei Xu, Jiaolong Yang, Jianfei Cai, Juyong Zhang, and Xin Tong. High-resolution optical flow from 1d attention and correlation. In ICCV, pages 10498–10507, 2021.

[19] Feihu Zhang, Oliver J. Woodford, Victor Adrian Prisacariu, and Philip H.S. Torr. Separable flow: Learning motion cost volumes for optical flow estimation. In ICCV, pages 10807–10817, October 2021.

[20] Huang, Zhaoyang, et al. "Flowformer: A transformer architecture for optical flow." European Conference on Computer Vision. Cham: Springer Nature Switzerland, 2022.

[21] Sui, Xiuchao, et al. "CRAFT: Cross-attentional flow transformer for robust optical flow." Proceedings of the IEEE/CVF conference on Computer Vision and Pattern Recognition. 2022.

[22] Dong, Qiaole, Chenjie Cao, and Yanwei Fu. "Rethinking Optical Flow from Geometric Matching Consistent Perspective." Proceedings of the IEEE/CVF Conference on Computer Vision and Pattern Recognition. 2023.

[23] Jung, Hyunyoung, et al. "AnyFlow: Arbitrary Scale Optical Flow with Implicit Neural Representation." Proceedings of the IEEE/CVF Conference on Computer Vision and Pattern Recognition. 2023.

[24] Shi, Xiaoyu, et al. "Flowformer++: Masked cost volume autoencoding for pretraining optical flow estimation." Proceedings of the IEEE/CVF Conference on Computer Vision and Pattern Recognition. 2023.

[25] Bai, Shaojie, et al. "Deep equilibrium optical flow estimation." Proceedings of the IEEE/CVF Conference on Computer Vision and Pattern Recognition. 2022.

[26] Simon Meister, Junhwa Hur, and Stefan Roth. Unflow: Unsupervised learning of optical flow with a bidirectional census loss. arXiv preprint arXiv:1711.07837, 2017.

[27] Pengpeng Liu, Michael Lyu, Irwin King, and Jia Xu. Selflow: Self-supervised learning of optical flow. In Proceedings of the IEEE Conference on Computer Vision and Pattern Recognition, pages 4571–4580, 2019.

[28] Shengyu Zhao, Yilun Sheng, Yue Dong, Eric I Chang, Yan Xu, et al. Maskflownet: Asymmetric feature matching with learnable occlusion mask. In Proceedings of the IEEE/CVF Conference on Computer Vision and Pattern Recognition, pages 6278–6287, 2020.

[29] Junhwa Hur and Stefan Roth. Iterative residual refinement for joint optical flow and occlusion estimation. In Proceedings of the IEEE Conference on Computer Vision and Pattern Recognition, pages 5754–5763, 2019.

[30] Rico Jonschkowski, Austin Stone, Jonathan T Barron, Ariel Gordon, Kurt Konolige, and Anelia Angelova. What matters in unsupervised optical flow. arXiv preprint arXiv:2006.04902, 2020.

[31] Jeong, Jisoo, et al. "Imposing consistency for optical flow estimation." Proceedings of the IEEE/CVF conference on Computer Vision and Pattern Recognition. 2022.

[32] Deng, Haowen, Tolga Birdal, and Slobodan Ilic. "Ppfnet: Global context aware local features for robust 3d point matching." Proceedings of the IEEE conference on computer vision and pattern recognition. 2018.

[33] Cao, Bingyi, Andre Araujo, and Jack Sim. "Unifying deep local and global features for image search." Computer Vision–ECCV 2020: 16th European Conference, Glasgow, UK, August 23–28, 2020, Proceedings, Part XX 16. Springer International Publishing, 2020.

[34] Ding, Yikang, et al. "Transmvsnet: Global context-aware multi-view stereo network with transformers." Proceedings of the IEEE/CVF Conference on Computer Vision and Pattern Recognition. 2022.

[35] Paszke, Adam, et al. "Pytorch: An imperative style, high-performance deep learning library." Advances in neural information processing systems 32 (2019)..